\documentclass[10pt,letterpaper,journal]{IEEEtran}
\def\TEVCAppendicesInMain{}
\ifdefined\pdfminorversion
\fi
\usepackage{amsmath,amssymb,bm}
\usepackage{booktabs,array,tabularx,longtable}
\usepackage{multirow}
\usepackage{graphicx,xcolor}
\usepackage{microtype}
\usepackage{float,placeins,needspace}
\floatstyle{ruled}
\newfloat{algorithm}{tbp}{loa}
\floatname{algorithm}{Algorithm}
\floatstyle{plain}
\usepackage{stfloats}
\usepackage{balance}
\usepackage{fvextra}
\usepackage{etoolbox}
\usepackage[nocompress]{cite}
\usepackage{url}
\usepackage{xr-hyper}
\usepackage[hidelinks]{hyperref}
\newcommand{\doi}[1]{\href{https://doi.org/#1}{\nolinkurl{#1}}}

\newcolumntype{Y}{>{\centering\arraybackslash}X}
\newcommand{\resultnum}[1]{\makebox[2.6em][r]{#1}}

\newcommand{\TEVCmissingfigure}[1]{%
  \fbox{\begin{minipage}[c][0.26\textheight][c]{0.92\linewidth}\centering
    \small Figure file \texttt{\detokenize{#1}} not found in the source package.\end{minipage}}}
\newcommand{\supplementaryprefix}{\ifdefined\TEVCAppendicesInMain\else Supplementary\space\fi}
\newcommand{\resulttableformat}{%
  \normalsize
  \setlength{\tabcolsep}{2.5pt}%
  \renewcommand{\arraystretch}{0.96}%
  \setlength{\aboverulesep}{0.3ex}%
  \setlength{\belowrulesep}{0.5ex}%
}
\makeatletter
\patchcmd{\@BAlancecol}{\unvbox1\vfil}
  {\dimen2=\dp1\unvbox1\kern-\dimen2\vfil}
  {}{\PackageError{TEVC}{Column balancing patch failed}{Check the balance package version.}}
\patchcmd{\@BAlancecol}{\global\advance\@tempdima by 1pt}
  {\global\advance\@tempdima by 0.01pt}{}{}
\ifdefined\TEVCStandaloneSupplement
  \patchcmd{\LT@output}{\copy\LT@foot\vss}{\copy\LT@foot\vfil}{}{}
  \patchcmd{\LT@output}{\copy\LT@foot\vss}{\copy\LT@foot\vfil}{}{}
\fi
\def\@IEEEabskeysecsize{\normalsize}
\let\TEVCoriginalmakecaption\@makecaption
\long\def\@makecaption#1#2{{\let\footnotesize\normalsize\TEVCoriginalmakecaption{#1}{#2}}}
\patchcmd{\thebibliography}{\footnotesize}{\normalsize}{}{}
\makeatother
\ifdefined\TEVCAppendicesInMain
\let\TEVCbodymonofont\ttdefault
\usepackage{inconsolata}
\let\TEVCappendixmonofont\ttdefault
\let\ttdefault\TEVCbodymonofont
\makeatletter
\long\def\LT@makecaption#1#2#3{%
  \LT@mcol\LT@cols c{\hbox to\z@{\hss
    \parbox[t]{\LTcapwidth}{\def\@captype{table}\@makecaption{#2}{#3}}%
    \hss}}}
\makeatother

\else
\fi
\hypersetup{pdftitle={AutoSECO: Automated Skill Evolution for Cross-Problem Combinatorial Optimization with General-Purpose LLM Agents},pdfauthor={},pdfsubject={Anonymous TEVC regular-paper submission}}
\title{AutoSECO: Automated Skill Evolution for Cross-Problem Combinatorial Optimization with General-Purpose LLM Agents}
\author{
Zirui Zhu, Cong Zhang, Ni Zhang, Kuan Xu, and Zhiguang Cao
\thanks{Zirui Zhu is with the Whiting School of Engineering, Johns Hopkins University, USA (e-mail: zzhu87@alumni.jh.edu)}
\thanks{Cong Zhang used to be affiliated with Nanyang Technological University, Huawei Noah’s Ark Lab, and TikTok Singapore. Now he is an independent researcher.}
\thanks{Ni Zhang, Kuan Xu and Zhiguang Cao are with the School of Computing and Information Systems, Singapore Management University, Singapore.}
}
\begin{document}
\bstctlcite{TEVCreferenceStyle}
\maketitle
\begin{abstract}
General-purpose large language model (LLM) agents can develop complete solvers for combinatorial optimization problems. Their specialization, however, often relies on workflows engineered for individual problems, limiting the reuse of development expertise. We propose AutoSECO, an automated skill evolution method that learns reusable solver-development guidance from experience. The challenge is that external evaluation measures the completed solver, while the skill guides decisions during development. AutoSECO connects this feedback to guidance deficiencies by tracing performance shortfalls through solver code and recorded development decisions. Revisions are expressed as general instructions and retained only if a fresh solver developed under the revised skill achieves a lower mean evaluation gap. Evolved solely on the traveling salesperson problem (TSP), the skill guides fresh agents in developing solvers for capacitated vehicle routing, job-shop scheduling, and three-dimensional bin packing without further skill evolution. On TSP, the final mean benchmark gap is 0.52\%, compared with 0.98\% for the skill-evolution baseline SkillOpt. Across all four problems, the evolved skill achieves better aggregate results than no skill guidance, the initial skill, and SkillOpt. The same frozen skill also supports solver development under alternative model--harness configurations. These findings suggest that solver-development experience can be retained as reusable expertise for general-purpose LLM agents.
\end{abstract}

%\begin{abstract}
%General-purpose LLM agents built on agent loops and execution harnesses can perform complex tasks, yet remain generalists. In combinatorial optimization, existing agentic systems often rely on manually engineered workflows tailored to individual problems, limiting reuse and scalability. This raises the question of how a general agent can acquire reusable expertise across heterogeneous problems. We propose AutoSECO, an automated skill evolution method that learns reusable behavioral guidance from solver-development experience. The method traces solver performance shortfalls through implemented designs and recorded development decisions to identify guidance to revise. Each revision is tested through fresh solver development and retained only when it improves solution quality. A skill evolved solely on the traveling salesperson problem transfers without further evolution to capacitated vehicle routing, job-shop scheduling, and three-dimensional bin packing. Across these unseen problems, the skill improves aggregate solution quality over no skill guidance, the initial skill, and a recent skill-evolution baseline. These findings suggest that solver-development experience can become reusable expertise, allowing general agents to improve across diverse combinatorial optimization problems.
%\end{abstract}

\section{Introduction}
\label{sec:intro}

\looseness=-1 % Avoid a short final line in this paragraph.
General-purpose agents built on large language models (LLMs) can reason about goals, use tools, execute code, and revise their actions in response to feedback~\cite{yao2023react,shinn2023reflexion,wang2024codeact,yang2024sweagent}. For combinatorial optimization problems (COPs), these capabilities support both the generation of heuristics or algorithmic components~\cite{romeraparedes2024funsearch,liu2024eoh,ye2024reevo,novikov2025alphaevolve,zhu2026simpleevol} and the development of complete, executable solvers~\cite{chen2026heurigym,sun2026cobench,xu2026dream}. This development process requires the agent to connect problem analysis and mechanism design with implementation, testing, and revision. Agentic optimization systems often supply expertise through specialized roles, staged workflows, or interfaces to dedicated solvers~\cite{ahmaditeshnizi2024optimus,yuan2026formuevo}. When expertise is embedded in system design, extending it to a different COP can require additional engineering. This motivates learning development guidance that a general-purpose LLM agent can reuse across heterogeneous COPs.

Skills offer a way to express such expertise as editable external guidance. In this work, a skill consists of natural-language instructions that the agent consults during solver development. Its effectiveness depends on how the agent interprets and applies those instructions, so manual design alone does not ensure the intended behavior. Even small wording changes can affect task performance~\cite{zhan2024lexicalsensitivity}. A skill should therefore be assessed and revised against the behavior and outcomes it produces. Recent skill evolution methods refine guidance by distilling trajectories, diagnosing failures, or selecting revisions through validation~\cite{ni2026trace2skill,wang2026skillgrad,yang2026skillopt}. In optimization, OptSkills learns reusable modeling-and-solving workflows~\cite{yang2026optskills}, and OptiSkill evolves a bank of skills~\cite{zhao2026optiskill}, both with the correctness of modeling as their primary objective. These studies provide a foundation for retaining experience as guidance, while complete solver development poses a particular challenge in interpreting the available feedback.

The difficulty is that feedback for revising a solving skill is indirect. External evaluation scores the completed solver, while the skill guides the decisions through which it is developed. A performance shortfall may reflect a limiting design, an implementation error, or insufficient guidance, and the score alone does not indicate whether or how the skill should change. Revising the guidance therefore requires examining the solver code alongside the observations and decisions recorded during development. This context helps identify where the guidance itself may be missing, ambiguous, or too general to support an effective development decision.

We propose Automated Skill Evolution for Combinatorial Optimization (AutoSECO), a method that turns this development evidence into reusable guidance for general-purpose agents. A solver agent develops a complete solver under the current \emph{solving skill}, while a \emph{skill evolver}, guided by a fixed \emph{evolver skill}, examines the resulting evidence. The evolver traces observed weaknesses through solver code and recorded development decisions to identify guidance that may need revision. The changes are expressed as general development instructions and tested by having a fresh agent build a new solver from scratch. A candidate replaces the current skill only if the new solver's mean evaluation gap is strictly lower than the gap previously recorded under the current skill. Only the solving skill evolves, while the agent configuration, evolver skill, and selection rule remain fixed.

Using the traveling salesperson problem (TSP) as the sole source COP, the final AutoSECO skill achieves a mean benchmark gap of 0.52\%, compared with 0.98\% for SkillOpt. The same skill is held fixed while fresh agents develop solvers for capacitated vehicle routing (CVRP), job-shop scheduling (JSSP), and three-dimensional bin packing (3D-BPP), using each target problem's task definition and training instances. Across all four COPs, AutoSECO achieves better aggregate results than No Skill, Initial Skill, and SkillOpt. The frozen skill also supports solver development under alternative model--harness configurations. These findings support solver-development experience as a source of reusable COP expertise, expressed through guidance that can be applied across problems and agent configurations.

\section{Related Work}
\label{sec:related}

\noindent\emph{a) Automated Solver Development.}
Automated heuristic design searches for effective rules, components, or combinations of heuristics, with production scheduling as a major application~\cite{branke2016schedulingreview}. Cross-domain reuse is also a goal of hyper-heuristics. For example, Sabar et al. evolve heuristic-selection and acceptance rules through gene expression programming and evaluate the framework across six COP domains~\cite{sabar2015gephh}. LLM-based approaches extend this search through program generation and execution feedback~\cite{wu2025llmecsurvey,romeraparedes2024funsearch,liu2024eoh,ye2024reevo,novikov2025alphaevolve,zhu2026directing}, including complete metaheuristics for continuous optimization in LLaMEA~\cite{vanstein2025llamea}. Adaptation can also occur in the process that generates code. EvoPH co-evolves guiding prompts and heuristics~\cite{liu2026evoph}, while LLaMoCo adapts a model through instruction tuning for optimization code generation~\cite{ma2026llamoco}.

Agentic frameworks organize the broader process of formulation, implementation, evaluation, and revision through combinations of specialized roles, workflows, and search procedures~\cite{zhang2026agentic,xu2026dream,li2025solverllm,zhang2026llm}. HeuriGym and CO-Bench assess these capabilities through executable heuristics and algorithms produced by agents~\cite{chen2026heurigym,sun2026cobench}. Within this broader development process, AutoSECO examines how external guidance for a fixed agent configuration can be refined through experience.

\noindent\emph{b) Skill Evolution.}
Skill evolution makes execution experience reusable by updating agent instructions and supporting resources. Trajectory analysis can consolidate local lessons into skills or persistent knowledge~\cite{ni2026trace2skill,tang2026wikiskill}, while failure diagnosis can provide directions for targeted edits~\cite{alzubi2026evoskill,wang2026skillgrad}. Validation and verification help assess which proposed updates to retain~\cite{yang2026skillopt,zhang2026coevoskills}. These mechanisms are complementary and can appear together within the same system.

SkillOpt, for example, combines success and failure analysis with bounded addition, replacement, and deletion edits and a validation gate~\cite{yang2026skillopt}. AutoSECO shares this reliance on execution evidence and validated updates. Its focus is on making the diagnostic link explicit for COP solver development: evaluation weaknesses are traced through the solver implementation and recorded development decisions to the relevant skill guidance. Each revision is tested through fresh solver development.

\noindent\emph{c) Experience and Skill Learning for Combinatorial Optimization.}
Optimization-modeling research uses solver feedback to turn solving attempts into reusable experience. AlphaOPT extracts solver-verified insights from failed modeling attempts and refines their applicability conditions~\cite{kong2025alphaopt}. Experience can also be organized around problem structure. OptSkills groups problems by archetype and distills successful trajectories into reusable workflows~\cite{yang2026optskills}, while OptiSkill maintains solver-verified modeling experience in a hierarchical, evolving SkillBank~\cite{zhao2026optiskill}. These studies emphasize correctly formulating and solving natural-language optimization problems, with modeling accuracy central to their evaluations. AutoSECO instead evaluates the solution quality of a developed solver across benchmark instances.

Related heuristic-design research also reuses experience through evolution strategies, accumulated mechanism-level knowledge, and explicit search over design hypotheses~\cite{yang2025heuragenix,yang2026ares,nguyen2026knowledgefirst}. These studies establish reusable knowledge as an important part of algorithm design. Building on this perspective, AutoSECO focuses on guidance for the agent's diagnosis, experimentation, and code revision during complete solver development. Its cross-problem evaluation tests whether guidance evolved on one COP remains useful when fresh agents develop solvers for other COPs.

\section{Method}
\label{sec:method}

%AutoSECO evolves a reusable solving skill for developing complete COP solvers with a general-purpose LLM agent (Fig.~\ref{fig:architecture}). A solver agent develops a solver under the current skill, and an external evaluator measures its performance. Guided by the evolver skill, the skill evolver traces performance shortfalls through solver code and recorded development decisions to identify instructions to revise. It proposes a candidate skill, under which a fresh solver agent develops a new solver. The candidate replaces the current skill if the new solver’s mean evaluation gap is lower than the current skill’s recorded gap. Only the solving skill evolves; the evolver skill and selection rule remain fixed.

\begin{figure*}[!t]
  \centering
  \IfFileExists{figures_skill_evolution.pdf}{\includegraphics[width=\textwidth]{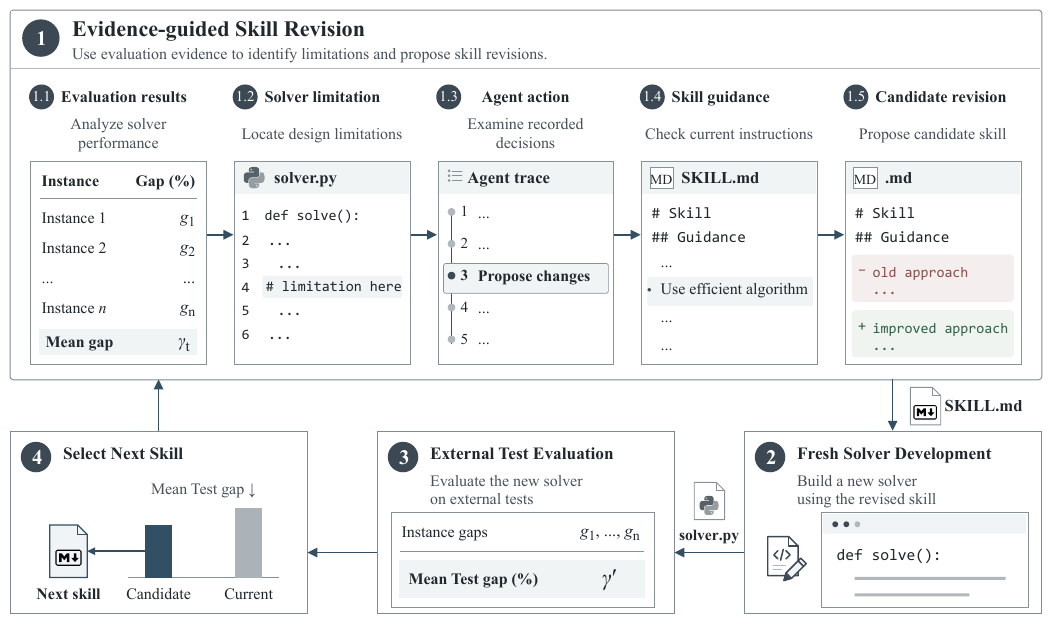}}{\TEVCmissingfigure{figures_skill_evolution.pdf}}
  \caption{Overview of AutoSECO. Evaluation feedback and development records guide skill revision. Each candidate is evaluated through fresh solver development and replaces the current skill if the mean evaluation gap decreases.}
  \label{fig:architecture}
\end{figure*}

AutoSECO learns reusable guidance for COP solver development by examining the decisions behind an agent’s solver and their implications for solution quality (Fig.~\ref{fig:architecture}). Solver code and development records provide the context needed to interpret performance shortfalls and identify where the current solving skill may offer insufficient guidance. These insights are expressed as general development instructions, allowing experience with one solver to inform how a general-purpose LLM agent approaches future solver-development tasks. Fresh solver development provides an empirical test of whether the revised guidance helps the agent build a better solver.

%\subsection{Solver Development under the Solving Skill}
%\label{sec:solver-development}
%The manually written initial skill \(S_0\) specifies a ReAct-style loop for solver development (Fig.~\ref{fig:solver-development}). Training instances are problem instances provided to the agent for testing and revising its solver. The agent implements a solver, runs it on these instances, examines the solver’s performance and search behavior, and revises the code before testing again. Across skill versions, the loop remains fixed while guidance for interpreting results and choosing code revisions evolves.
%Solver implementations are restricted to Python’s standard library, NumPy~\cite{harris2020numpy}, Numba~\cite{numba_docs}, and basic SciPy numerical functions~\cite{2020SciPy-NMeth}. The agent has no access to web search or external solvers during development.
%\input{figures_en_solver_development}

\subsection{Solver Development under the Solving Skill}
\label{sec:solver-development}

\begin{figure*}[t]
  \centering
  \IfFileExists{figures_solver_development.pdf}{\includegraphics[width=\textwidth]{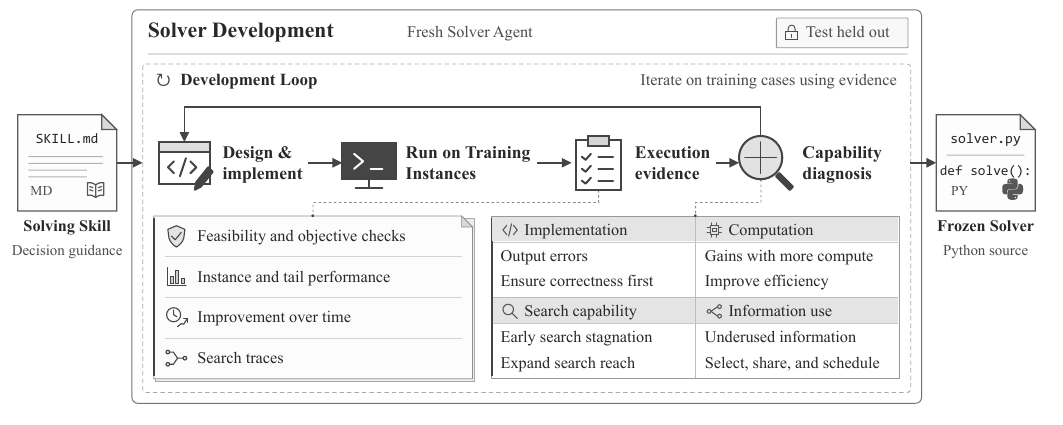}}{\TEVCmissingfigure{figures_solver_development.pdf}}
  \caption{Skill-guided solver development with the test set held out. Under a fixed skill, a fresh agent uses execution evidence from training instances to diagnose capability gaps and iteratively revise solver code.}
  \label{fig:solver-development}
\end{figure*}

\looseness=-1 % Avoid a short final line in this paragraph.
The solving skill guides the agent in diagnosing solver limitations and deciding how to address them (Fig.~\ref{fig:solver-development}). The manually written initial skill \(S_0\) organizes this process as a ReAct-style development loop, in which the agent uses solver performance and search behavior on training instances to assess and refine its implementation. Within each development run, the skill remains fixed as the agent revises the solver code. Evolution preserves this development loop while refining the guidance for interpreting execution evidence and choosing code revisions.

The development evidence concerns both implementation correctness and the use of computational and search capacity. Feasibility and objective checks indicate whether the current solver satisfies the task contract. Performance across training instances, improvement over time, and search traces reveal where it remains weak and how additional computation is used. Continued improvement with low evaluation throughput motivates examining efficiency, while early stagnation despite more computation motivates examining search reach. Here, search reach refers to the moves or solution structures explored within the solver's runtime budget. The guidance also addresses whether available information informs selection, sharing, and scheduling, for example when search statistics are used to allocate time across search channels. A search channel is a separately budgeted search process. These observations support code revisions within a run and provide the context for subsequent skill revision.

Solver implementations are restricted to Python's standard library, NumPy~\cite{harris2020numpy}, Numba~\cite{numba_docs}, and basic SciPy numerical functions~\cite{2020SciPy-NMeth}. The agent has no access to web search or external solvers during development.

\subsection{From Evaluation Feedback to Skill Revision}
\label{sec:method-diagnosis}
\label{sec:method-edit}

\looseness=-1 % Avoid a short final line in this paragraph.
For the current skill \(S_t\) in round \(t\), the skill evolver uses external evaluation results to investigate where the guidance may provide insufficient support for solver development. Guided by the fixed evolver skill, it interprets instance-level results together with the completed solver code and the agent's development records. These sources answer different questions: the results locate the weakness, the code shows what was implemented, and the record reveals the observations and explanations behind its adoption. The trace links these sources around a development decision that the skill was intended to guide.

Patterns in the evaluation results help focus the analysis on weaknesses associated with particular problem scales or structures, recurring across rounds, or concentrated in a tail of high-gap instances. The evolver examines the solver code for design limitations consistent with these patterns and uses the development records to locate the decisions that introduced or retained the relevant design. The observations available to the agent at those points and its recorded explanations provide the context for assessing the corresponding guidance in the current solving skill \(S_t\).

This trace provides an explanation consistent with the recorded evidence. It does not isolate the causal effect of an instruction, because a solver reflects interacting development choices. Its purpose is to identify a plausible revision target for empirical assessment.

A revision requires evidence that the guidance itself is insufficient, for example because an instruction is missing, ambiguous, conflicting, or too general to inform the decision. A failure to follow clear instructions, an isolated coding error, or an incomparable run does not by itself justify changing the skill. When the evidence does not support a skill-level explanation, the evolver leaves that issue untouched.

\looseness=-1 % Avoid a short final line in this paragraph.
The resulting revision expresses a general lesson about solver development in terms of problem structure, training observations, and resource limits. It may clarify an instruction, resolve conflicting advice, or introduce a missing diagnostic consideration. The revision is framed around observations and choices available in a future development run. This allows a new agent to apply the lesson using its own task definition and training evidence, without access to the source solver's benchmark results.

For example, low evaluation throughput accompanied by continued improvement on training instances may motivate clearer guidance to check computational efficiency before replacing the search mechanism. The source solver informs the revision, but the guidance does not reproduce its particular algorithm or encode benchmark identities or external evaluation outcomes. The revised skill becomes a candidate \(C_t\), whose value is assessed through fresh solver development in Section~\ref{sec:method-selection}.

\subsection{Skill Selection by Fresh Solver Performance}
\label{sec:method-selection}

A candidate skill \(C_t\) is assessed by the quality of the solver a fresh agent develops under its guidance. The agent develops this solver from scratch in a single run, using the same COP task, training instances, and development budget as those used for the retained skill \(S_t\). The candidate skill remains fixed while the agent tests and revises the solver code.

A fresh run tests whether the revised guidance remains useful when the agent must make the implementation choices again. The retained solver code and its development record are not carried into this run, so the candidate is assessed through a new development trajectory.

The completed solver is externally evaluated on the same instances used for \(S_t\). Selection depends solely on whether its mean gap is strictly lower than the previously recorded mean gap of the solver developed under \(S_t\), with both values compared before rounding. The solver under the retained skill is not rebuilt for this comparison.

Let \(\operatorname{Score}(S)\) denote the negative mean evaluation gap recorded for the solver developed under skill \(S\). A higher score indicates a lower observed gap, giving the update rule
\begin{equation}
S_{t+1}=
\begin{cases}
C_t, & \operatorname{Score}(C_t) > \operatorname{Score}(S_t),\\
S_t, & \operatorname{Score}(C_t) \leq \operatorname{Score}(S_t).
\end{cases}
\label{eq:skill-selection}
\end{equation}
An accepted candidate's measured gap becomes the reference for the next round. Otherwise, the retained skill and its recorded gap remain unchanged.

%\subsection{Skill Selection by Fresh Solver Performance}
%\label{sec:method-selection}

%Each candidate skill \(C_t\) is assessed in one development run with a fresh solver agent.
%The agent develops a solver from scratch using the same COP task, training instances, and development budget as those used for the retained skill \(S_t\).
%The candidate skill remains fixed during this run, while the agent implements, tests, and revises the solver code.

%An external evaluator runs the completed solver on the same evaluation instances used for \(S_t\) and computes its mean gap. We compare this value with the previously recorded mean gap of the solver developed under \(S_t\). The comparison uses unrounded values, and \(C_t\) replaces \(S_t\) only if the new mean gap is strictly lower. Acceptance depends solely on this comparison of mean gaps.

%Let $\operatorname{Score}(S)$ denote the negative mean evaluation gap recorded for the solver developed under skill $S$. A higher score therefore corresponds to a lower observed mean gap. The update rule is

%\begin{equation}
%S_{t+1}=
%\begin{cases}
%C_t, & \operatorname{Score}(C_t) > \operatorname{Score}(S_t),\\
%S_t, & \operatorname{Score}(C_t) \leq \operatorname{Score}(S_t).
%\end{cases}
%\label{eq:skill-selection}
%\end{equation}
%If \(C_t\) is accepted, its measured mean gap becomes the reference for the next round. A tie or a higher mean gap leaves \(S_t\) and its recorded gap unchanged.

\subsection{Overall Skill-Evolution Procedure}
\label{sec:algorithm}

Algorithm~\ref{alg:skill-evolution} maintains a solving skill together with its associated development evidence. At the start of round \(t\), the retained skill \(S_t\) is accompanied by its solver \(F_t\), development record \(H_t\), instance-level evaluation results \(G_t\), and recorded mean gap \(\gamma_t\). These records provide the basis for diagnosing limitations and proposing the next revision. Within each round, primes denote the candidate's solver \(F'\), development record \(H'\), evaluation results \(G'\), and mean gap \(\gamma'\). The scores in Eq.~(\ref{eq:skill-selection}) satisfy \(\operatorname{Score}(S_t)=-\gamma_t\) and \(\operatorname{Score}(C_t)=-\gamma'\).

External evaluation uses instances \(D_{\mathrm{test}}\), which are withheld from the solver agent during development. Their instance-level results are provided to the skill evolver for diagnosis, while their mean gap determines candidate selection. The evolver skill, selection rule, COP task, instance sets, and development budget remain fixed throughout evolution.

\begin{algorithm}[!t]
\caption{Solving-skill evolution in AutoSECO.}
\label{alg:skill-evolution}
\normalsize
\noindent\textbf{Input:} initial solving skill $S_0$, fixed evolver skill $E$, COP task $P$, training instances $D_{\mathrm{tr}}$, evaluation instances $D_{\mathrm{test}}$, development time budget $B$, and number of candidate rounds $R$.\par
\noindent\textbf{Output:} final solving skill $S_R$.\par\smallskip
\setlength{\tabcolsep}{3pt}
\renewcommand{\arraystretch}{1.02}
\begin{tabularx}{\linewidth}{@{}r>{\raggedright\arraybackslash}X@{}}
1 & With a fresh agent, develop $F_0$ from scratch under fixed $S_0$, using $P$, $D_{\mathrm{tr}}$, and $B$, and record $H_0$. \\
2 & Evaluate $F_0$ on $D_{\mathrm{test}}$ to obtain instance-level results $G_0$ and mean gap $\gamma_0$. \\
3 & \textbf{for} $t=0,\ldots,R-1$ \textbf{do} \\
4 & \hspace{0.5em} Following $E$, trace a performance shortfall in $G_t$ through $F_t$ and $H_t$ to assess the relevant guidance in $S_t$. \\
5 & \hspace{0.5em} Propose $C_t$ from revisions to the general development guidance in $S_t$ that are supported by the evidence. \\
6 & \hspace{0.5em} In one run, develop $F'$ from scratch with a fresh agent under fixed $C_t$, using $P$, $D_{\mathrm{tr}}$, and $B$, and record $H'$. \\
7 & \hspace{0.5em} Evaluate $F'$ on $D_{\mathrm{test}}$ to obtain instance-level results $G'$ and mean gap $\gamma'$. \\
8 & \hspace{0.5em} \textbf{if} $\gamma'<\gamma_t$ (using unrounded gaps) \textbf{then} \\
9 & \quad $(S_{t+1},F_{t+1},H_{t+1},G_{t+1},\gamma_{t+1})\gets(C_t,F',H',G',\gamma')$. \\
10 & \hspace{0.5em} \textbf{else} \\
11 & \quad $(S_{t+1},F_{t+1},H_{t+1},G_{t+1},\gamma_{t+1})\gets(S_t,F_t,H_t,G_t,\gamma_t)$. \\
12 & \hspace{0.5em} \textbf{end if} \\
13 & \textbf{end for} \\
14 & \textbf{return} $S_R$. \\
\end{tabularx}
\end{algorithm}

Acceptance carries forward the candidate skill together with its solver, development record, and evaluation results. Rejection preserves the existing state, so subsequent revisions continue to use evidence associated with the retained skill.

The strict acceptance rule makes the retained skill's recorded mean gap non-increasing across rounds. Because each candidate is assessed through a single stochastic development run, this property does not establish monotonic improvement in expected solver quality. Final benchmark results are averaged over three fresh development runs with the selected skill \(S_R\) fixed.

\section{Experiments and Results}
\label{sec:results}

%We first compare AutoSECO and SkillOpt on TSP to examine how solver quality changes over successive evolution rounds. We then evaluate cross-problem transfer to CVRP, JSSP, and 3D-BPP, holding the final evolved skills fixed while fresh agents develop solvers for each problem. Finally, we deploy the skill evolved by AutoSECO, without further evolution, under alternative model–harness configurations and evaluate the resulting solvers.

\looseness=-1 % Avoid a short final line in this paragraph.
The experiments assess whether AutoSECO improves the quality of solvers developed by general-purpose agents and produces guidance that remains useful beyond its source setting. Comparisons with SkillOpt on TSP examine solver quality over successive evolution rounds. The evolved skills are held fixed while fresh agents develop solvers for CVRP, JSSP, and 3D-BPP, testing whether guidance acquired on TSP transfers to heterogeneous COPs. The AutoSECO skill is also deployed without further evolution under alternative model--harness configurations to assess its usefulness across different executors.

\subsection{Experimental Setup}
\label{sec:experiments}
\subsubsection{Models and Software}
\label{sec:runtime-environment}

Primary experiments use GPT-5.6 Sol~\cite{openai2026gpt56} with the Codex harness~\cite{openai2026codex}. To assess transfer across executors, the AutoSECO skill evolved under this configuration is deployed with two alternative pairings: GPT-5.6 Sol with Pi~\cite{pi2026codingagent}, and DeepSeek-V4-Flash~\cite{deepseek2026v4} with Codex. The skill remains fixed throughout these deployment experiments. Table~\ref{tab:runtime-environment} lists the model and software versions.

\begin{table}[!htbp]
  \centering
  \caption{Models and software versions.}
  \label{tab:runtime-environment}
  \normalsize
  \setlength{\tabcolsep}{5pt}
  \renewcommand{\arraystretch}{1.02}
  \begin{tabularx}{\columnwidth}{@{}>{\raggedright\arraybackslash}p{0.43\columnwidth} >{\raggedright\arraybackslash}X@{}}
    \toprule
    \textbf{Component} & \textbf{Version} \\
    \midrule
    Main model & GPT-5.6 Sol \\
    Alternative model & DeepSeek-V4-Flash \\
    \addlinespace[2pt]
    Agent harness & Codex 0.155.0-alpha.9.2 \\
    Alternative harness & Pi 0.85.1 \\
    \addlinespace[2pt]
    Language runtime & Python 3.13.9 \\
    NumPy & 2.3.5 \\
    Numba & 0.62.1 \\
    SciPy & 1.16.3 \\
    \bottomrule
  \end{tabularx}
\end{table}

Experiments run on an Intel Core i7-13650HX CPU with 16~GB RAM under Windows~11. Solver development has a four-hour wall-clock budget per run, and solver execution is limited to eight threads.

\subsubsection{Datasets and Evaluation Protocol}
\label{sec:data-splits}

Solver agents develop and refine their code using synthetic training instances that are disjoint from the corresponding public benchmarks (Table~\ref{tab:dataset-composition}). Along with the task definition, agents receive the training instances, their objective targets, and optional reference solutions. Benchmark instances and their fixed reference values are withheld from solver agents.

\begin{table}[!t]
  \centering
  \caption{Training scales and public benchmarks.}
  \label{tab:dataset-composition}
  \normalsize
  \setlength{\tabcolsep}{0pt}
  \renewcommand{\arraystretch}{1.02}
  \begin{tabularx}{\columnwidth}{@{}l@{\hspace{7pt}}>{\raggedright\arraybackslash}p{0.405\columnwidth}@{\hspace{8pt}}>{\raggedright\arraybackslash}X@{}}
    \toprule
    \textbf{COP} & \textbf{Training} & \textbf{Evaluation benchmark} \\
    \midrule
    TSP
      & Six node sizes:\newline 50, 100, 200, 500,\newline 1,000, 10,000.
      & TSPLIB EUC\_2D~\cite{reinelt1991tsplib}\newline 77 instances;\newline 51--18,512 nodes. \\
    \addlinespace[4pt]
    CVRP
      & Five customer sizes: 50,\newline 100, 200, 500, 1,000.
      & CVRPLIB-X~\cite{uchoa2017cvrp}\newline 100 instances. \\
    \addlinespace[4pt]
    JSSP
      & Multiple sizes of\newline jobs and machines.
      & Taillard and DMU\newline 80 each~\cite{taillard1993benchmarks,demirkol1998benchmarks};\newline $15\!\times\!15$--$100\!\times\!20$. \\
    \addlinespace[4pt]
    3D-BPP
      & Eight MPV classes; 50,\newline 100, 150, or 200 items.
      & MPV320; eight classes\newline 320 instances~\cite{martello2000threedimensional,goncalves2013brkga}. \\
    \bottomrule
  \end{tabularx}
\end{table}

The TSP benchmark serves as the evolution and selection set \(D_{\mathrm{test}}\), with instance-level evaluation feedback available to the skill evolver. In the cross-problem experiments, the skills evolved on TSP remain fixed while fresh agents develop solvers for CVRP, JSSP, and 3D-BPP. The completed solvers are evaluated on the corresponding public benchmarks.

Development runs are isolated, and agents have no access to web search or external solvers such as LKH. During development, agents can evaluate solvers on training instances in parallel, whereas benchmark evaluation proceeds sequentially across instances. The \emph{No Skill} condition uses the same task inputs, tools, and development budget as the skill-guided conditions, with the skill text omitted.

\subsubsection{Task Definitions and Runtime Limits}
\label{sec:problem-contracts}

The external evaluator checks solution feasibility and recomputes the objective value. Generic task-definition, configuration, and prompt templates are provided in the \emph{\nameref{app:execution-templates}} section of the \supplementaryprefix{}Appendix. The following runtime limits apply to individual instances during final benchmark evaluation.

\paragraph{TSP}
The task is about the 2D travelling salesman problem in the Euclidean space. A solution is a permutation of node IDs, and the evaluator verifies that each node appears exactly once before computing the closed-tour length using TSPLIB integer-rounded Euclidean distances. Runtime limits range from 10s for $n\le100$ to 150s for $n>10{,}000$, with intermediate tiers of 15, 30, 45, 60, 90, and 120s.

\paragraph{CVRP}
The task uses one depot, identical vehicle capacities, an unlimited fleet, and integer-rounded Euclidean distances. A solution specifies a nonempty customer sequence for each route. The evaluator verifies that every customer is served exactly once and that route capacities are respected, then computes the total route length. Runtime limits range from 20s for $n<200$ to 150s for $n=1{,}000$, with intermediate tiers of 30, 45, and 90s.

\paragraph{JSSP}
Each job visits every machine once, with positive integer processing times. A solution assigns a nonnegative integer start time to each operation. The evaluator checks job precedence and machine non-overlap before computing the makespan. Runtime limits are 60s for $15\!\times\!15$ and $20\!\times\!15$, 180s for $100\!\times\!20$, and 120s otherwise.

\paragraph{3D-BPP}
Items have fixed orientations and are packed into an unlimited supply of identical bins. A solution assigns each item to a consecutively numbered bin at integer coordinates. The evaluator checks complete assignment, containment, and non-overlap before counting the bins used. For an instance with $n$ items, the runtime limit is \mbox{$T(n)=60+120((n-50)/150)^{1.4}$} seconds.

\subsubsection{Metrics and Repeated Runs}
\label{sec:evaluation-metrics}

Reference values are fixed before experimentation. TSP uses proven optima, and CVRP uses best-known upper bounds (BKS). For JSSP and 3D-BPP (i.e., MPV320), instances with proven optima are termed \emph{closed}, while those evaluated against fixed BKS values are termed \emph{open}. JSSP contains 103 closed and 57 open instances, and MPV320 contains 21 closed and 299 open instances.

For TSP, CVRP, and JSSP, the instance-level gap is
\begin{equation}
\label{eq:instance-gap}
g(x)=\frac{z(x)-\bar z(x)}{\bar z(x)},
\end{equation}
where $z(x)$ is the objective value recomputed by the evaluator and $\bar z(x)$ is the fixed reference value. We report $100g(x)$ as the percentage gap. Negative gaps indicate improvements over a fixed BKS value.

\looseness=-1 % Avoid a short final line in this paragraph.
Each candidate skill is assessed through one fresh solver-development run during evolution. Final results for the four skill conditions and the alternative model--harness configurations are averaged over three fresh solver-development runs for each COP. Within each run, the mean gap for TSP, CVRP, and JSSP weights benchmark instances equally. For 3D-BPP, we compute the mean bin count within each class and the total bin count over all 320 instances, then average these quantities across the three runs.

\begin{figure*}[t]
  \centering
  \IfFileExists{figures_evolution_gap.pdf}{\includegraphics[width=\textwidth]{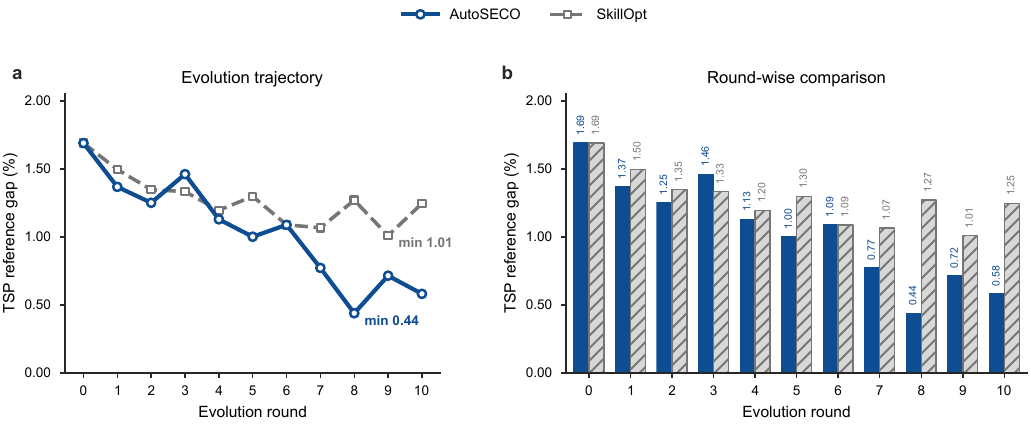}}{\TEVCmissingfigure{figures_evolution_gap.pdf}}
  \caption{TSP benchmark mean gap of each round's candidate solver. The curves include rejected candidates and show candidate performance rather than the sequence of retained skills.}
  \label{fig:evolution-gap}
\end{figure*}

\subsection{Skill Evolution on TSP}
\label{sec:evolution-results}
AutoSECO reaches a lower best recorded mean TSP gap than SkillOpt~\cite{yang2026skillopt} over ten evolution rounds (Fig.~\ref{fig:evolution-gap}). Both methods start from the same Initial Skill with a baseline gap of 1.69\% and use GPT-5.6 Sol with the Codex harness. AutoSECO reaches its lowest recorded gap of 0.44\% in round~8, while SkillOpt reaches 1.01\% in round~9. The skills selected in these rounds remain retained through the end of evolution under the selection rule in Section~\ref{sec:method-selection}.

\looseness=-1 % Avoid a short final line in this paragraph.
The figure reports the initial assessment as round~0 and successive candidate assessments as rounds~1--10. Each result comes from a single fresh solver-development run. Rejected candidates remain visible in the curves, which therefore show candidate performance rather than the sequence of retained skills. \supplementaryprefix{}Tables~\ref{tab:evolution-ours} and~\ref{tab:evolution-skillopt} summarize the revisions, candidate results, and selection decisions for all rounds.

\iffalse
\subsection{Skill Evolution on TSP}
\label{sec:evolution-results}
AutoSECO and SkillOpt start from the same Initial Skill and complete ten evolution rounds using GPT-5.6 Sol with the Codex harness. Each candidate skill is evaluated through one fresh solver-development run, and the skill retained by the selection gate is used in the next round (Section~\ref{sec:method-selection}). Fig.~\ref{fig:evolution-gap} shows the mean TSP benchmark gaps of the initial and candidate solvers, including those associated with rejected skill revisions.

The best recorded mean gaps are 0.44\% for AutoSECO (round~8) and 1.01\% for SkillOpt (round~9), down from a shared baseline of 1.69\%. The skills selected in AutoSECO round~8 and SkillOpt round~9 are retained through the end of evolution. \supplementaryprefix{}Tables~\ref{tab:evolution-ours} and~\ref{tab:evolution-skillopt} provide the revisions, candidate results, and selection decisions for all rounds.
\fi

\subsection{Comparison of Evolution Processes and Evolved Skills}
\label{sec:evolution-analysis}
\begin{table*}[!tbp]
  \centering
  \caption{Solver-development guidance in the initial and evolved skills.}
  \label{tab:skill-guidance-comparison}
  \normalsize
  \setlength{\tabcolsep}{6pt}
  \renewcommand{\arraystretch}{1.02}
  \begin{tabularx}{\linewidth}{@{}>{\raggedright\arraybackslash}p{0.15\linewidth}*{3}{>{\raggedright\arraybackslash}X}@{}}
    \toprule
    \textbf{Development decision} & \textbf{Initial Skill} & \textbf{SkillOpt} & \textbf{AutoSECO} \\
    \midrule
    \textbf{Performance diagnosis}
      & Inspect instance features; identify poorly performing groups.
      & Group instances by computable features; prioritize recurring weak groups.
      & Link computable features to required search capabilities; test the link with counterexamples. \\
    \addlinespace[5pt]
    \textbf{Stagnation diagnosis}
      & Distinguish slow evaluation with continued improvement from early search stagnation.
      & Examine improvement curves and neighborhood success rates.
      & Compare paired runs across budgets; distinguish low evaluation throughput from limited search reach. \\
    \addlinespace[5pt]
    \textbf{Mechanism validation}
      & Specify which moves change; test components separately.
      & Compare performance with and without the mechanism on weak groups under equal budgets.
      & Verify distinct effective moves in a prototype; test feature-diverse instances and the full~training~set. \\
    \addlinespace[5pt]
    \textbf{Search-time allocation}
      & Record runtime, candidate counts, and improvements for each search channel.
      & Allocate time by recent improvement per unit time; retain exploration.
      & Consider recent and long-term improvement per unit time; preserve channels that reach different solution structures. \\
    \bottomrule
  \end{tabularx}
  \par\smallskip
  \begin{minipage}{\linewidth}
    \normalsize
    \textit{Note:} The SkillOpt and AutoSECO columns summarize the final solving skills evolved by the respective methods. A search channel is a separately budgeted search process.
  \end{minipage}
\end{table*}

\paragraph{Revision targeting}
Both methods use execution feedback to revise skills and retain candidates with lower mean gaps. They differ in how this feedback identifies guidance worth revising. AutoSECO traces evaluation weaknesses through solver design and recorded development decisions to the relevant skill instructions. The SkillOpt~\cite{yang2026skillopt} procedure used in this comparison derives local additions, replacements, and deletions from recurring failure and success patterns in agent trajectories, without explicitly requiring this trace.

From the end of round~6 to the end of round~8, the retained mean gap decreases from 1.002\% to 0.440\% for AutoSECO and from 1.090\% to 1.068\% for SkillOpt. During this interval, AutoSECO's accepted revisions address search-time allocation and validation of distinct effective moves. SkillOpt revises fixed-threshold tests in round~7, while its round~8 proposal to remove full training-set validation is rejected. \supplementaryprefix{}Tables~\ref{tab:evolution-ours} and~\ref{tab:evolution-skillopt} document the revisions and selection outcomes.

\paragraph{Retained skill guidance}
Table~\ref{tab:skill-guidance-comparison} compares how the Initial Skill and the two final evolved skills support the same solver-development decisions. The Initial Skill already distinguishes slow evaluation with continued improvement from early search stagnation. SkillOpt uses improvement curves and neighborhood success rates to support this diagnosis. AutoSECO specifies paired runs across runtime budgets to distinguish low evaluation throughput from limited search reach, as defined in Section~\ref{sec:solver-development}. It also requires staged tests of whether new mechanisms add effective moves and considers both recent and long-term improvement per unit time when allocating search budgets. The full skill texts are available from the repository listed in the \emph{\nameref{app:skill-artifacts}} section of the \supplementaryprefix{}Appendix.

\paragraph{Accepted and rejected revisions}
Rounds~2, 8, and~9 in \supplementaryprefix{}Table~\ref{tab:evolution-ours} provide examples of recorded revision targets and candidate selection outcomes.

In round~2, low solution-evaluation throughput is observed alongside continued improvement late in the runtime budget. This pattern motivates guidance to probe throughput when improvement persists and to probe search reach when early stagnation persists despite additional computation of the same kind. The candidate solver obtains a mean gap of 1.25\%, below the retained 1.37\%, and the revision is accepted. The resulting guidance expresses a diagnostic distinction that can be applied beyond the source solver's implementation.

In round~8, the search channels still reach a similar structural plateau after the budget-allocation revision in round~7. The candidate skill requires evidence that an additional mechanism enables distinct effective moves, assessed through a prototype, a feature-diverse panel, and full training-set evaluation. Its solver achieves a mean gap of 0.44\%, compared with the retained 0.77\%, and the revision is accepted.

Round~9 illustrates why a proposed instruction still requires evaluation through fresh solver development. The candidate skill recommends reserving budget to strengthen an online lower bound when it indicates remaining headroom. The associated record describes lower-bound computation consuming search time without reliably improving decisions or final quality. The candidate solver obtains a mean gap of 0.72\%, above the retained 0.44\%, and the revision is rejected.

These examples connect solver evidence, proposed guidance, and selection outcomes. Each outcome concerns a complete candidate skill assessed through one fresh development run and does not isolate the contribution of an individual instruction.

\subsection{Final Skill Evaluation Across COPs}
\label{sec:main-results}

Table~\ref{tab:main-results} compares No Skill, Initial Skill, SkillOpt, and AutoSECO across four COPs. SkillOpt and AutoSECO use the final skills selected after evolution on TSP. For each COP and condition, fresh agents develop solvers from scratch using that problem's task definition and training instances. All supplied skills remain fixed during development. Results for these four conditions are averaged over three fresh solver-development runs for each COP.

\begin{table*}[!t]
  \centering
  \caption{Solver performance across four combinatorial optimization problems.}
  \label{tab:main-results}
  \resulttableformat
  \newcommand{\bppbinsize}[2]{%
    \multicolumn{2}{@{}l}{\makebox[0.15\textwidth][l]{#1}\makebox[0.075\textwidth][l]{#2}}}
  \begin{tabularx}{\textwidth}{@{}p{0.105\textwidth}p{0.12\textwidth}YYYYYY@{}}
    \toprule
    \multicolumn{8}{@{}l}{\textbf{TSP --- TSPLIB EUC\_2D} (mean gap \%; node scale)} \\
    \textbf{Benchmark} & \textbf{Scale} & \textbf{No Skill} & \textbf{Initial} & \textbf{SkillOpt} & \textbf{AutoSECO} & \textbf{ReEvo} & \textbf{DRAGON} \\
    \midrule
    TSPLIB & 50--500 & \resultnum{2.14} & \resultnum{\underline{0.07}} & \resultnum{0.25} & \resultnum{\textbf{0.02}} & \resultnum{8.17} & \resultnum{9.74} \\
     & 501--1,000 & \resultnum{3.30} & \resultnum{0.96} & \resultnum{\underline{0.73}} & \resultnum{\textbf{0.36}} & \resultnum{16.61} & \resultnum{11.24} \\
     & 1,001--2,000 & \resultnum{4.19} & \resultnum{1.74} & \resultnum{\underline{1.29}} & \resultnum{\textbf{0.63}} & \resultnum{18.23} & \resultnum{15.24} \\
     & 2,001--5,000 & \resultnum{5.05} & \resultnum{2.72} & \resultnum{\underline{2.34}} & \resultnum{\textbf{1.52}} & \resultnum{21.20} & \resultnum{19.37} \\
     & 5,001--20,000 & \resultnum{11.48} & \resultnum{10.89} & \resultnum{\underline{3.50}} & \resultnum{\textbf{2.43}} & \resultnum{18.62} & \resultnum{15.42} \\
    \cmidrule(lr){3-8}
    \multicolumn{2}{@{}l}{\textit{Average}} & \resultnum{3.75} & \resultnum{1.69} & \resultnum{\underline{0.98}} & \resultnum{\textbf{0.52}} & \resultnum{13.10} & \resultnum{12.24} \\

    \midrule
    \multicolumn{8}{@{}l}{\textbf{CVRP --- CVRPLIB-X} (mean gap \%; customer scale)} \\
    \textbf{Benchmark} & \textbf{Scale} & \textbf{No Skill} & \textbf{Initial} & \textbf{SkillOpt} & \textbf{AutoSECO} & \textbf{ReEvo} & \textbf{AFL} \\
    \midrule
    X & 100--200 & \resultnum{7.58} & \resultnum{3.59} & \resultnum{2.62} & \resultnum{\textbf{0.52}} & \resultnum{8.77} & \resultnum{\underline{1.93}} \\
     & 201--500 & \resultnum{6.41} & \resultnum{4.79} & \resultnum{\underline{3.83}} & \resultnum{\textbf{1.33}} & \resultnum{14.88} & \resultnum{5.20} \\
     & 501--1,000 & \resultnum{6.45} & \resultnum{4.68} & \resultnum{\underline{4.04}} & \resultnum{\textbf{2.00}} & \resultnum{19.81} & \resultnum{6.66} \\
    \cmidrule(lr){3-8}
    \multicolumn{2}{@{}l}{\textit{Average}} & \resultnum{6.68} & \resultnum{4.49} & \resultnum{\underline{3.63}} & \resultnum{\textbf{1.37}} & \resultnum{15.11} & \resultnum{4.95} \\

    \midrule
    \multicolumn{8}{@{}l}{\textbf{JSSP --- Taillard and DMU} (mean gap to fixed reference \%; jobs $\times$ machines)} \\
    \textbf{Family} & \textbf{Shape} & \textbf{No Skill} & \textbf{Initial} & \textbf{SkillOpt} & \textbf{AutoSECO} & \textbf{NDS-S} & \multicolumn{1}{c@{}}{\mbox{\textbf{SLIM (GM$_{512}$)}}} \\
    \midrule
    Taillard & $15\!\times\!15$ & \resultnum{18.84} & \resultnum{3.09} & \resultnum{\underline{2.75}} & \resultnum{\textbf{0.84}} & \resultnum{3.20} & \resultnum{6.50} \\
     & $20\!\times\!15$ & \resultnum{22.97} & \resultnum{5.56} & \resultnum{\underline{4.78}} & \resultnum{\textbf{2.22}} & \resultnum{5.50} & \resultnum{8.80} \\
     & $20\!\times\!20$ & \resultnum{20.54} & \resultnum{5.14} & \resultnum{\underline{4.29}} & \resultnum{\textbf{1.86}} & \resultnum{5.20} & \resultnum{9.00} \\
     & $30\!\times\!15$ & \resultnum{23.16} & \resultnum{6.32} & \resultnum{\underline{4.53}} & \resultnum{\textbf{2.20}} & \resultnum{7.40} & \resultnum{10.60} \\
     & $30\!\times\!20$ & \resultnum{28.07} & \resultnum{9.10} & \resultnum{\underline{7.45}} & \resultnum{\textbf{3.91}} & \resultnum{10.80} & \resultnum{12.70} \\
     & $50\!\times\!15$ & \resultnum{17.47} & \resultnum{6.48} & \resultnum{\underline{0.48}} & \resultnum{\textbf{0.03}} & \resultnum{8.10} & \resultnum{4.90} \\
     & $50\!\times\!20$ & \resultnum{19.88} & \resultnum{8.06} & \resultnum{\underline{2.60}} & \resultnum{\textbf{0.53}} & \resultnum{10.20} & \resultnum{7.60} \\
     & $100\!\times\!20$ & \resultnum{9.82} & \resultnum{5.13} & \resultnum{\underline{0.12}} & \resultnum{\textbf{0.00}} & \resultnum{9.70} & \resultnum{2.10} \\
    \addlinespace[2pt]
    DMU & $20\!\times\!15$ & \resultnum{27.95} & \resultnum{8.69} & \resultnum{\underline{7.01}} & \resultnum{\textbf{3.63}} & \resultnum{8.60} & \resultnum{11.30} \\
     & $20\!\times\!20$ & \resultnum{27.19} & \resultnum{8.46} & \resultnum{\underline{7.18}} & \resultnum{\textbf{3.47}} & \resultnum{9.20} & \resultnum{12.30} \\
     & $30\!\times\!15$ & \resultnum{30.61} & \resultnum{15.36} & \resultnum{\underline{8.37}} & \resultnum{\textbf{4.76}} & \resultnum{17.10} & \resultnum{14.00} \\
     & $30\!\times\!20$ & \resultnum{31.98} & \resultnum{15.17} & \resultnum{\underline{8.44}} & \resultnum{\textbf{7.09}} & \resultnum{17.20} & \resultnum{15.80} \\
     & $40\!\times\!15$ & \resultnum{26.76} & \resultnum{15.38} & \resultnum{\underline{10.28}} & \resultnum{\textbf{4.70}} & \resultnum{18.80} & \resultnum{10.90} \\
     & $40\!\times\!20$ & \resultnum{33.36} & \resultnum{18.67} & \resultnum{\underline{13.26}} & \resultnum{\textbf{6.90}} & \resultnum{20.50} & \resultnum{14.80} \\
     & $50\!\times\!15$ & \resultnum{24.96} & \resultnum{18.63} & \resultnum{12.84} & \resultnum{\textbf{5.45}} & \resultnum{24.90} & \resultnum{\underline{10.60}} \\
     & $50\!\times\!20$ & \resultnum{31.25} & \resultnum{21.59} & \resultnum{16.38} & \resultnum{\textbf{8.03}} & \resultnum{25.10} & \resultnum{\underline{15.00}} \\
    \cmidrule(lr){3-8}
    \multicolumn{2}{@{}l}{\textit{Average}} & \resultnum{24.67} & \resultnum{10.68} & \resultnum{\underline{6.92}} & \resultnum{\textbf{3.48}} & \resultnum{12.59} & \resultnum{10.45} \\

    \midrule
    \multicolumn{8}{@{}l}{\textbf{3D-BPP --- MPV320} (mean bins; final row: mean total bins $\downarrow$)} \\
    \bppbinsize{\textbf{Bin size}}{\textbf{Class}} & \textbf{No Skill} & \textbf{Initial} & \textbf{SkillOpt} & \textbf{AutoSECO} & \textbf{BRKGA-BPP} & \textbf{HGA-NR} \\
    \midrule
    \bppbinsize{$100\!\times\!100\!\times\!100$}{1} & \resultnum{33.00} & \resultnum{32.70} & \resultnum{32.12} & \resultnum{31.88} & \resultnum{\underline{31.80}} & \resultnum{\textbf{31.58}} \\
    \bppbinsize{$100\!\times\!100\!\times\!100$}{2} & \resultnum{32.65} & \resultnum{32.23} & \resultnum{31.78} & \resultnum{\underline{31.39}} & \resultnum{\textbf{31.35}} & \resultnum{31.43} \\
    \bppbinsize{$100\!\times\!100\!\times\!100$}{3} & \resultnum{32.95} & \resultnum{32.54} & \resultnum{32.01} & \resultnum{\underline{31.65}} & \resultnum{\textbf{31.63}} & \resultnum{\underline{31.65}} \\
    \bppbinsize{$100\!\times\!100\!\times\!100$}{4} & \resultnum{73.53} & \resultnum{\underline{73.50}} & \resultnum{73.61} & \resultnum{\underline{73.50}} & \resultnum{\underline{73.50}} & \resultnum{\textbf{73.33}} \\
    \bppbinsize{$100\!\times\!100\!\times\!100$}{5} & \resultnum{18.20} & \resultnum{18.17} & \resultnum{17.87} & \resultnum{\underline{17.60}} & \resultnum{\underline{17.60}} & \resultnum{\textbf{17.43}} \\
    \bppbinsize{$10\!\times\!10\!\times\!10$}{6} & \resultnum{24.23} & \resultnum{24.17} & \resultnum{24.03} & \resultnum{\underline{23.76}} & \resultnum{\textbf{23.73}} & \resultnum{23.90} \\
    \bppbinsize{$40\!\times\!40\!\times\!40$}{7} & \resultnum{15.63} & \resultnum{15.24} & \resultnum{14.87} & \resultnum{\underline{14.59}} & \resultnum{\textbf{14.58}} & \resultnum{\textbf{14.58}} \\
    \bppbinsize{$100\!\times\!100\!\times\!100$}{8} & \resultnum{21.38} & \resultnum{21.09} & \resultnum{20.63} & \resultnum{\underline{20.29}} & \resultnum{\textbf{20.25}} & \resultnum{20.35} \\
    \cmidrule(lr){3-8}
    \multicolumn{2}{@{}l}{\textit{Mean total bins}} &
    10062.0 & 9985.3 & 9875.7 & 9786.3 &
    \underline{9777} & \textbf{9769} \\
    \bottomrule
  \end{tabularx}
  \par\smallskip
  \begin{minipage}{\textwidth}
    \normalsize\raggedright
    \textit{Note:} Initial denotes Initial Skill; SkillOpt and AutoSECO use their final evolved skills. Lower values are better. The best result in each row is shown in bold, and the second best is underlined.
  \end{minipage}
\end{table*}

% Queue the deployment table early so it does not interrupt the discussion.
\begin{table*}[!t]
  \centering
  \caption{Solver performance across model--harness configurations.}
  \label{tab:model-agent}
  \label{tab:model-agent-detailed}
  \resulttableformat
  \newcommand{\deploymentbinsize}[2]{\makebox[\linewidth][s]{#1\hfill #2}}
  \renewcommand{\arraystretch}{0.96}
  \setlength{\aboverulesep}{0.25ex}
  \setlength{\belowrulesep}{0.4ex}
  \setlength{\tabcolsep}{3pt}
  % Extra space only between JSSP body rows; keep both headers identical.
  \newcommand{\deploymentjsspgap}{2.445pt}

  %==================== Left panel ====================
  \begin{minipage}[t]{0.48\textwidth}
    \vspace{0pt}
    \centering
    \begin{tabularx}{\linewidth}{@{}p{0.3\linewidth}*{3}{>{\hsize=0.95\hsize\linewidth=\hsize\centering\arraybackslash}X}>{\hsize=1.15\hsize\linewidth=\hsize\centering\arraybackslash}X@{}}
      \toprule
      \multirow{4}{*}{\textbf{Scale / Class}} & \multirow{4}{*}{\shortstack{\textbf{Initial}\\\textbf{Skill}}} & \multicolumn{3}{c}{\textbf{AutoSECO}} \\
      \cmidrule(l){3-5}
       & & \multicolumn{2}{c}{\shortstack{\textbf{GPT-5.6}\\\textbf{Sol}}} & \shortstack{\textbf{DeepSeek}\\\textbf{V4-Flash}} \\
      \cmidrule(lr){3-4}\cmidrule(l){5-5}
       & & Codex & Pi & Codex \\
      \midrule
      \multicolumn{5}{@{}l}{\textbf{TSP --- TSPLIB EUC\_2D (nodes)}} \\
      50--500 & \resultnum{0.07} & \resultnum{0.02} & \resultnum{0.02} & \resultnum{0.09} \\
      501--1,000 & \resultnum{0.96} & \resultnum{0.36} & \resultnum{0.36} & \resultnum{0.53} \\
      1,001--2,000 & \resultnum{1.74} & \resultnum{0.63} & \resultnum{0.63} & \resultnum{0.94} \\
      2,001--5,000 & \resultnum{2.72} & \resultnum{1.52} & \resultnum{1.37} & \resultnum{1.54} \\
      5,001--20,000 & \resultnum{10.89} & \resultnum{2.43} & \resultnum{2.75} & \resultnum{2.39} \\
      \cmidrule(l){2-5}
      \textit{Average} & \resultnum{1.69} & \resultnum{0.52} & \resultnum{0.54} & \resultnum{0.63} \\
      \addlinespace[3pt]
      \multicolumn{5}{@{}l}{\textbf{CVRP --- CVRPLIB-X (customers)}} \\
      100--200 & \resultnum{3.59} & \resultnum{0.52} & \resultnum{0.81} & \resultnum{1.92} \\
      201--500 & \resultnum{4.79} & \resultnum{1.33} & \resultnum{1.64} & \resultnum{2.89} \\
      501--1,000 & \resultnum{4.68} & \resultnum{2.00} & \resultnum{2.59} & \resultnum{3.46} \\
      \cmidrule(l){2-5}
      \textit{Average} & \resultnum{4.49} & \resultnum{1.37} & \resultnum{1.76} & \resultnum{2.86} \\
      \addlinespace[3pt]
      \multicolumn{5}{@{}l}{\textbf{3D-BPP --- MPV320 (bin size; class)}} \\
      \deploymentbinsize{$100\!\times\!100\!\times\!100$}{1} & \resultnum{32.70} & \resultnum{31.88} & \resultnum{31.89} & \resultnum{32.37} \\
      \deploymentbinsize{$100\!\times\!100\!\times\!100$}{2} & \resultnum{32.23} & \resultnum{31.39} & \resultnum{31.37} & \resultnum{32.01} \\
      \deploymentbinsize{$100\!\times\!100\!\times\!100$}{3} & \resultnum{32.54} & \resultnum{31.65} & \resultnum{31.66} & \resultnum{32.23} \\
      \deploymentbinsize{$100\!\times\!100\!\times\!100$}{4} & \resultnum{73.50} & \resultnum{73.50} & \resultnum{73.50} & \resultnum{73.50} \\
      \deploymentbinsize{$100\!\times\!100\!\times\!100$}{5} & \resultnum{18.17} & \resultnum{17.60} & \resultnum{17.60} & \resultnum{17.94} \\
      \deploymentbinsize{$10\!\times\!10\!\times\!10$}{6} & \resultnum{24.17} & \resultnum{23.76} & \resultnum{23.76} & \resultnum{24.06} \\
      \deploymentbinsize{$40\!\times\!40\!\times\!40$}{7} & \resultnum{15.24} & \resultnum{14.59} & \resultnum{14.62} & \resultnum{14.93} \\
      \deploymentbinsize{$100\!\times\!100\!\times\!100$}{8} & \resultnum{21.09} & \resultnum{20.29} & \resultnum{20.30} & \resultnum{20.71} \\
      \cmidrule(l){2-5}
      \textit{Mean total bins} &
9985.3 & 9786.3 & 9787.7 & 9909.7 \\
      \bottomrule
    \end{tabularx}
  \end{minipage}\hfill
  %==================== Right panel ====================
  \begin{minipage}[t]{0.48\textwidth}
    \vspace{0pt}
    \centering
    \begin{tabularx}{\linewidth}{@{}p{0.24\linewidth}*{3}{>{\hsize=0.95\hsize\linewidth=\hsize\centering\arraybackslash}X}>{\hsize=1.15\hsize\linewidth=\hsize\centering\arraybackslash}X@{}}
      \toprule
      \multirow{4}{*}{\textbf{Scale / Class}} & \multirow{4}{*}{\shortstack{\textbf{Initial}\\\textbf{Skill}}} & \multicolumn{3}{c}{\textbf{AutoSECO}} \\
      \cmidrule(l){3-5}
       & & \multicolumn{2}{c}{\shortstack{\textbf{GPT-5.6}\\\textbf{Sol}}} & \shortstack{\textbf{DeepSeek}\\\textbf{V4-Flash}} \\
      \cmidrule(lr){3-4}\cmidrule(l){5-5}
       & & Codex & Pi & Codex \\
      \midrule
      \multicolumn{5}{@{}l}{\textbf{JSSP --- Taillard (jobs $\times$ machines)}} \\[\deploymentjsspgap]
      $15\!\times\!15$ & \resultnum{3.09} & \resultnum{0.84} & \resultnum{1.15} & \resultnum{1.15} \\[\deploymentjsspgap]
      $20\!\times\!15$ & \resultnum{5.56} & \resultnum{2.22} & \resultnum{2.66} & \resultnum{3.36} \\[\deploymentjsspgap]
      $20\!\times\!20$ & \resultnum{5.14} & \resultnum{1.86} & \resultnum{2.60} & \resultnum{2.94} \\[\deploymentjsspgap]
      $30\!\times\!15$ & \resultnum{6.32} & \resultnum{2.20} & \resultnum{2.84} & \resultnum{3.04} \\[\deploymentjsspgap]
      $30\!\times\!20$ & \resultnum{9.10} & \resultnum{3.91} & \resultnum{5.22} & \resultnum{5.82} \\[\deploymentjsspgap]
      $50\!\times\!15$ & \resultnum{6.48} & \resultnum{0.03} & \resultnum{0.04} & \resultnum{0.17} \\[\deploymentjsspgap]
      $50\!\times\!20$ & \resultnum{8.06} & \resultnum{0.53} & \resultnum{1.40} & \resultnum{1.89} \\[\deploymentjsspgap]
      $100\!\times\!20$ & \resultnum{5.13} & \resultnum{0.00} & \resultnum{0.04} & \resultnum{0.48} \\[\deploymentjsspgap]
      \addlinespace[3pt]
      \multicolumn{5}{@{}l}{\textbf{JSSP --- DMU (jobs $\times$ machines)}} \\[\deploymentjsspgap]
      $20\!\times\!15$ & \resultnum{8.69} & \resultnum{3.63} & \resultnum{5.17} & \resultnum{5.47} \\[\deploymentjsspgap]
      $20\!\times\!20$ & \resultnum{8.46} & \resultnum{3.47} & \resultnum{6.30} & \resultnum{4.88} \\[\deploymentjsspgap]
      $30\!\times\!15$ & \resultnum{15.36} & \resultnum{4.76} & \resultnum{7.60} & \resultnum{8.19} \\[\deploymentjsspgap]
      $30\!\times\!20$ & \resultnum{15.17} & \resultnum{7.09} & \resultnum{7.98} & \resultnum{9.73} \\[\deploymentjsspgap]
      $40\!\times\!15$ & \resultnum{15.38} & \resultnum{4.70} & \resultnum{6.56} & \resultnum{8.17} \\[\deploymentjsspgap]
      $40\!\times\!20$ & \resultnum{18.67} & \resultnum{6.90} & \resultnum{8.29} & \resultnum{11.63} \\[\deploymentjsspgap]
      $50\!\times\!15$ & \resultnum{18.63} & \resultnum{5.45} & \resultnum{6.30} & \resultnum{10.24} \\[\deploymentjsspgap]
      $50\!\times\!20$ & \resultnum{21.59} & \resultnum{8.03} & \resultnum{8.75} & \resultnum{13.79} \\[\deploymentjsspgap]
      \cmidrule(l){2-5}
      \textit{JSSP average} & \resultnum{10.68} & \resultnum{3.48} & \resultnum{4.56} & \resultnum{5.68} \\
      \bottomrule
    \end{tabularx}
  \end{minipage}

  \par\smallskip
  \begin{minipage}{\textwidth}
    \normalsize\raggedright
    \textit{Note:} Initial Skill uses GPT-5.6 Sol--Codex; all AutoSECO configurations use the same skill evolved on TSP and held fixed. TSP, CVRP, and JSSP: mean gap (\%); 3D-BPP: mean bins (last row: mean total bins). Lower is better.
  \end{minipage}
\end{table*}

Among the four conditions, AutoSECO achieves the best aggregate result on every COP, including the three problems not used for skill evolution. Relative to SkillOpt, it reduces the mean gap by 47\% on TSP (0.52\% vs.\ 0.98\%), 62\% on CVRP (1.37\% vs.\ 3.63\%), and 50\% on JSSP (3.48\% vs.\ 6.92\%). On 3D-BPP, it uses 89.4 fewer bins in total (9786.3 vs.\ 9875.7). Relative to the Initial Skill, mean gaps decrease by approximately 67--69\% on TSP, CVRP, and JSSP, while the total bin count on 3D-BPP decreases by 199.0.

The improvements extend across the reported instance groups. AutoSECO obtains the best or joint best result among the four conditions in all 32 groups: five TSP size groups, three CVRP size groups, sixteen JSSP groups, and eight 3D-BPP classes. The only tie is in 3D-BPP Class~4, where AutoSECO and the Initial Skill average 73.50 bins per instance.

The remaining gaps vary with instance scale and benchmark family. From the smallest to the largest reported size groups, AutoSECO's mean gap increases from 0.02\% to 2.43\% on TSP and from 0.52\% to 2.00\% on CVRP. On JSSP, the gap rounds to 0.00\% for Taillard $100\!\times\!20$ and is 0.03\% for Taillard $50\!\times\!15$, while DMU $50\!\times\!20$ retains a gap of 8.03\%. Thus, the aggregate improvements coexist with substantial differences in the difficulty of individual groups.

The additional baselines in Table~\ref{tab:main-results} provide comparisons with other approaches to solver design. On TSP, AutoSECO's mean gap of 0.52\% compares with 13.10\% for ReEvo~\cite{ye2024reevo} and 12.24\% for DRAGON~\cite{chen2026dragon}. On CVRP, its gap is 1.37\%, compared with 15.11\% for ReEvo and 4.95\% for AFL~\cite{zhang2026agentic}. On JSSP, the resulting solvers achieve a mean gap of 3.48\%, compared with 12.59\% for NDS-S~\cite{zago2026ndss} and 10.45\% for SLIM ($\mathrm{GM}_{512}$)~\cite{corsini2024slim}.

On 3D-BPP, the solvers developed under the TSP-evolved skill also approach the reported performance of specialized packing algorithms. Their mean total of 9786.3 bins exceeds the totals for HGA-NR and BRKGA-BPP~\cite{mahvash2018column,goncalves2013brkga,feng2015hybrid} by 17.3 and 9.3 bins, respectively, over all 320 instances.

\subsection{Deployment Across Model--Harness Configurations}
\label{sec:deployment-results}

Table~\ref{tab:model-agent} evaluates the same frozen AutoSECO skill under three model--harness configurations. The skill is evolved on TSP using GPT-5.6 Sol with Codex and deployed, without further evolution, with either GPT-5.6 Sol and Pi or DeepSeek-V4-Flash and Codex. Results are averaged over three fresh solver-development runs for each COP and configuration.

With GPT-5.6 Sol, replacing Codex with Pi preserves similar aggregate performance on TSP and 3D-BPP. The mean TSP gap changes from 0.52\% to 0.54\%, and the mean total bin count changes from 9786.3 to 9787.7. Larger differences appear on CVRP and JSSP, where mean gaps increase from 1.37\% to 1.76\% and from 3.48\% to 4.56\%, respectively.

Replacing GPT-5.6 Sol with DeepSeek-V4-Flash while retaining Codex yields higher aggregate gaps on all three gap-based benchmarks. The mean gaps are 0.63\% on TSP, 2.86\% on CVRP, and 5.68\% on JSSP. The mean total bin count on 3D-BPP increases to 9909.7.

Both alternative configurations nevertheless achieve better aggregate results on all four COPs than the common Initial Skill reference, which was evaluated using GPT-5.6 Sol with Codex. These results support reuse of the frozen skill across executors, although solver quality varies across configurations.

\section{Discussion}
\label{sec:discussion}

\noindent\textbf{Cross-Problem Generalization.}
The TSP-evolved skill improves aggregate solution quality on CVRP, JSSP, and 3D-BPP without further evolution (Table~\ref{tab:main-results}). Its guidance on diagnosis, mechanism testing, and resource allocation is reused while fresh agents develop new solvers using each target problem's task definition and training instances. This transfer suggests that development guidance can retain useful COP expertise.

\noindent\textbf{Evolution Dynamics.}
The Initial Skill already specifies the development loop, and evolution refines the guidance used within it. Paired budget comparisons help distinguish low evaluation throughput from limited search reach, while staged tests check whether a proposed mechanism adds distinct effective moves (Table~\ref{tab:skill-guidance-comparison}). The revisions summarized in \supplementaryprefix{}Table~\ref{tab:evolution-ours} illustrate how solver observations can motivate general development checks.

\noindent\textbf{Cross-Executor Deployment.}
\looseness=-1 % Avoid a short final line in this paragraph.
The frozen skill also supports solver development under alternative executors (Table~\ref{tab:model-agent}). In aggregate, the Pi configuration remains closer to the source results than the DeepSeek-V4-Flash configuration. This variation supports assessing the skill together with its model and harness~\cite{li2026skillsbench}. All three configurations achieve better aggregate results than the common Initial Skill reference, evaluated only with GPT-5.6 Sol and Codex. Matched baselines are needed to measure gains from skill evolution within each alternative configuration.

\noindent\textbf{Limitations.}
The study uses one source COP and fixed training sets, leaving the effects of source diversity, training coverage, and difficulty untested. For TSP, the same benchmark informs revision and selection and supplies the final scores, leaving generalization to new TSP instances untested. Each candidate is assessed through one stochastic development run, while final results average three fresh runs. Development variability can affect both selection and reported performance. Controlled ablations are also needed to separate the contributions of backward tracing and individual instructions.

\noindent\textbf{Future Directions.}
An open question is how guidance affects the decision to continue development or stop. Recording the evidence used at this point, together with the gains and costs of further development, could support stopping rules that balance expected improvement against additional effort. Evolution across multiple source COPs would further test whether broader development experience yields more transferable guidance.

\section{Conclusion}
\label{sec:conclusion}

AutoSECO evolves reusable guidance for developing complete COP solvers with general-purpose LLM agents. It traces performance shortfalls through solver code and recorded development decisions to identify guidance to revise, and assesses revisions through fresh solver development. The skill evolved solely on TSP achieves better aggregate results than No Skill, Initial Skill, and SkillOpt across TSP, CVRP, JSSP, and 3D-BPP, without further skill evolution on the target problems. The same frozen skill also supports solver development under alternative model--harness configurations. These findings suggest that solver-development experience can be retained as reusable expertise for general-purpose agents working across heterogeneous COPs.

%\section{Conclusion}
%\label{sec:conclusion}

%This paper presents AutoSECO, an automated skill evolution method for developing complete COP solvers with a general-purpose agent. AutoSECO links an observed weakness in solver performance to three evidence sources: the solver code, the agent's development decisions, and the current skill guidance. It expresses revisions as general development instructions and selects among them by fresh solver performance. The skill evolved by AutoSECO solely on TSP improves solver quality on TSP and transfers to CVRP, JSSP, and 3D-BPP. Across these tasks, this skill outperforms No Skill, Initial Skill, and SkillOpt in the overall benchmark results. The same frozen skill also remains effective under additional model--harness configurations. These findings support solver-development experience as a source of reusable COP expertise. Specifically, this experience is retained in guidance on diagnosing weaknesses, choosing experiments, and validating new search mechanisms. This guidance can be applied while the agent chooses the representation, feasibility checks, and search mechanisms required by a new task. In this way, automated skill evolution offers a practical way for general-purpose agents to build COP expertise from their solver-development experience.

% References stay directly after the text. Their last page is balanced only in
% the separate-document edition: when the appendix follows, balancing here
% interferes with the \onecolumn switch and overfills the output box.
\nobalance
\bibliographystyle{IEEEtran}
\bibliography{reference_style,references}
\ifdefined\TEVCAppendicesInMain
\else
  \makeatletter
  \if@firstcolumn\balance\fi
  \makeatother
\fi

% Single-PDF edition: the appendices follow the references. They are set
% one-column because the evolution tables are full-width longtables, and
% nothing follows them, so the document ends in one-column mode.
\ifdefined\TEVCAppendicesInMain
  \clearpage
  \nobalance
  \onecolumn
  \begingroup
  \let\ttdefault\TEVCappendixmonofont
  \fvset{baselinestretch=0.95}
  \appendix
  \section*{Round-by-Round Skill Evolution}
\label{app:evolution}

Tables~\ref{tab:evolution-ours} and~\ref{tab:evolution-skillopt} summarize the TSP evolution runs shown in Fig.~\ref{fig:evolution-gap}. Both methods use the same 77 TSPLIB instances for evaluation feedback and candidate selection. Round~0 reports the initial skill assessment. Rounds~1--10 correspond to the ten candidate proposals, indexed by \(t=0,\ldots,9\) in Algorithm~\ref{alg:skill-evolution}. Each reported gap comes from one fresh solver-development run, including candidates that are subsequently rejected.

Both runs start at 1.69\%. AutoSECO accepts revisions in rounds~1, 2, 4, 5, 7, and~8, and SkillOpt accepts revisions in rounds~1, 2, 3, 4, 6, 7, and~9. Rejection leaves the retained skill and its recorded gap unchanged. The qualitative columns summarize solver observations, design context, and revision targets. They do not reproduce complete development traces or isolate the effect of an individual instruction.

\subsection*{Skill Evolution with AutoSECO on TSP}
\label{app:evolution-ours}
\begingroup
\normalsize
\setlength{\tabcolsep}{2.4pt}
\renewcommand{\arraystretch}{1.05}
\setlength{\LTpre}{5pt}
\setlength{\LTpost}{6pt}
\begin{longtable}{@{}c
  >{\raggedright\arraybackslash}p{0.12\linewidth}
  >{\raggedright\arraybackslash}p{0.22\linewidth}
  >{\raggedright\arraybackslash}p{0.27\linewidth}
  >{\raggedright\arraybackslash}p{0.28\linewidth}@{}}
  \caption{AutoSECO evolution on TSP. Round 0 is the initial baseline. Later rows report single-run candidate mean gaps. Values are displayed to two decimals, while selection uses unrounded gaps.}
  \label{tab:evolution-ours}\\
  \toprule
  \textbf{Round} & \textbf{Gap and decision} & \textbf{Solver observations and design context} & \textbf{Recorded decisions and targets} & \textbf{Proposed skill revision} \\
  \midrule
  \endfirsthead

  \multicolumn{5}{l}{\normalsize Table~\ref{tab:evolution-ours} continued}\\
  \toprule
  \textbf{Round} & \textbf{Gap and decision} & \textbf{Solver observations and design context} & \textbf{Recorded decisions and targets} & \textbf{Proposed skill revision} \\
  \midrule
  \endhead

  \midrule
  \multicolumn{5}{r}{\normalsize Continued on next page}\\
  \endfoot

  \bottomrule
  \endlastfoot

  0 & 1.69\%; baseline retained
    & Feasible construction and small local moves stagnate early in strongly coupled regions; the same search workflow serves all instance types.
    & The agent extends local search after inspecting sizes and quality. Initial guidance lists features but does not require a testable feature-to-capability link.
    & No edit; Initial Skill establishes the baseline. \\

  1 & 1.37\%; accepted
    & Highly coupled regions inflate tail gap; one construction path produces similar weak starting solutions.
    & The agent relates input features to search needs. The skill does not yet require evidence for feature--mechanism fit or stress tests of valid feature combinations.
    & Require a computable feature, its implied capability, predicted benefit region, and counterexamples; check batch statistics and valid stress cases. \\

  2 & 1.25\%; accepted
    & Search continues to improve late in the budget, while solution-evaluation throughput remains low.
    & The agent compares short and standard-budget trajectories. The skill distinguishes computation from search qualitatively but gives no explicit diagnostic probe.
    & Test throughput first when late improvement persists; test search reach when early stagnation survives more computation of the same kind. \\

  3 & 1.46\%; rejected
    & Additional low-cost channels have different starts but heavily overlapping effective moves, consuming exploration budget.
    & The agent treats different starting paths as complementarity. The skill does not yet require novelty and return per CPU unit to establish channel value.
    & Try lightweight multi-channel exploration under uncertainty. \\

  4 & 1.13\%; accepted
    & More of the same local search reaches a quality plateau in coupled cases; partial teardown and reconstruction adds joint structural moves.
    & The agent compares extended search with short reconstruction probes. The skill allows larger neighborhoods but does not require a falsifiable capability hypothesis.
    & State trigger signals, newly reachable moves, expected benefit, and disconfirming outcomes; begin on a small feature panel. \\

  5 & 1.00\%; accepted
    & Mean improvement can conceal persistent weak regions; feature-conditioned construction and reconstruction budgets target them.
    & The agent compares paired runs by feature region and tail quality. The guidance does not make stable weak-region gains and absence of comparable new regressions explicit.
    & Promote an internal solver version only with preserved correctness, reliable target-region gain, interpretable contribution, and no equally severe new tail loss. \\

  6 & 1.09\%; rejected
    & A hard feature threshold routes reconstruction unstably near its boundary and in sparsely covered regions.
    & The agent turns a correlation into a hard switch without sufficient counterexamples. The skill warns against narrow thresholds but lacks a boundary-test rule.
    & Fit a normalized hard boundary and test both sides and unseen scales. \\

  7 & 0.77\%; accepted
    & Fixed budgets overfund low-return channels and risk neglecting complementary channels with delayed gains.
    & The agent uses online CPU, feasible-candidate, gain, and novelty signals. The guidance lacks an operational return measure and exploration protection.
    & Allocate by gain per CPU unit, novelty, and stagnation duration; do not eliminate a unique channel solely for short-term zero return. \\

  8 & 0.44\%; accepted
    & Despite improved scheduling, existing channels still reach the same structural plateau; a structured large neighborhood or local subproblem adds different moves.
    & The agent checks candidate coverage, improvement structure, and return before integration. The skill does not require evidence of distinct reachable moves.
    & Test reach with a prototype, then a feature-diverse panel, then full evaluation; reject mechanisms that repeat existing effective moves. \\

  9 & 0.72\%; rejected
    & A stronger lower bound consumes search time without reliably changing decisions or improving the final solution.
    & The agent treats better diagnostic information as a quality mechanism. The skill lacks an explicit test of the bound's runtime value.
    & Reserve budget to strengthen an online lower bound when it indicates headroom. \\

  10 & 0.58\%; rejected
    & A costly large neighborhood shows no gain in a short window. Such a window may miss rare large improvements.
    & The agent uses recent return alone. The skill lacks a sufficient test for distinguishing temporary silence from long-term ineffectiveness.
    & Reduce a channel to minimum exploration after one unproductive window. \\
\end{longtable}
\endgroup

\ifdefined\TEVCStandaloneSupplement
  \clearpage
\else
  \Needspace{8\baselineskip}
\fi
\subsection*{Skill Evolution with SkillOpt on TSP}
\label{app:evolution-skillopt}

Table~\ref{tab:evolution-skillopt} summarizes the local edits proposed during the SkillOpt run. Input skill identifies the version used to propose the edit, and the decision column identifies the skill carried forward. Counts \(a\rightarrow b\) refer to sentences in the edited passage before and after revision, with the net change in parentheses.

\begingroup
\normalsize
\setlength{\tabcolsep}{2.2pt}
\renewcommand{\arraystretch}{1.06}
\setlength{\LTpre}{5pt}
\setlength{\LTpost}{6pt}
\begin{longtable}{@{}c c
  >{\raggedright\arraybackslash}p{0.10\linewidth}
  >{\raggedright\arraybackslash}p{0.14\linewidth}
  >{\raggedright\arraybackslash}p{0.29\linewidth}
  >{\raggedright\arraybackslash}p{0.31\linewidth}@{}}
  \caption{SkillOpt evolution on TSP. Round 0 is the initial baseline. Later rows report single-run candidate mean gaps. Values are displayed to two decimals, while selection uses unrounded gaps.}
  \label{tab:evolution-skillopt}\\
  \toprule
  \textbf{Round} & \shortstack{\textbf{Input}\\\textbf{skill}} & \textbf{Gap} & \textbf{Decision and retained skill} & \textbf{Local edit} & \textbf{Motivation and observations} \\
  \midrule
  \endfirsthead

  \multicolumn{6}{l}{\normalsize Table~\ref{tab:evolution-skillopt} continued}\\
  \toprule
  \textbf{Round} & \shortstack{\textbf{Input}\\\textbf{skill}} & \textbf{Gap} & \textbf{Decision and retained skill} & \textbf{Local edit} & \textbf{Motivation and observations} \\
  \midrule
  \endhead

  \midrule
  \multicolumn{6}{r}{\normalsize Continued on next page}\\
  \endfoot

  \bottomrule
  \endlastfoot

  0 & --- & 1.69\% & Baseline; V0
    & No candidate edit.
    & The Initial Skill establishes the common baseline. \\

  1 & V0 & 1.50\% & Accept; V1
    & Addition, $0\rightarrow1$ ($+1$): group training instances by computable size and structural features and track recurring weak regions.
    & Overall quality can hide stable weak regions; grouped analysis identifies where improvement is needed. \\

  2 & V1 & 1.35\% & Accept; V2
    & Replacement, $1\rightarrow2$ ($+1$): diagnose early stagnation with improvement curves and neighborhood success before comparing search-scope mechanisms.
    & Extending the same search may not resolve stagnation; diagnosis should precede a mechanism change. \\

  3 & V2 & 1.33\% & Accept; V3
    & Addition, $0\rightarrow2$ ($+2$): predict target feature regions and use matched-budget on/off comparisons.
    & Concurrent mechanism and parameter changes make small gains hard to attribute; predictions and ablations sharpen the comparison. \\

  4 & V3 & 1.20\% & Accept; V4
    & Replacement, $1\rightarrow1$ (0): allocate channel budgets by recent effective gain per unit time while preserving weak-region exploration.
    & Static resource splits may limit effective throughput. Marginal gains can inform budget allocation. \\

  5 & V4 & 1.30\% & Reject; V4
    & Addition, $0\rightarrow1$ ($+1$): give the leading channel a fixed main budget and minimize all others.
    & An average lead need not generalize across weak regions. Concentrating the budget may exclude complementary search. \\

  6 & V4 & 1.09\% & Accept; V6
    & Addition, $0\rightarrow1$ ($+1$): require paired retests of persistent weak regions under the same budget.
    & Overall averages can conceal continued regional regression, which needs a separate check. \\

  7 & V6 & 1.07\% & Accept; V7
    & Replacement, $1\rightarrow1$ (0): normalize necessary fixed thresholds and test nearby values across scales.
    & A fixed threshold can change behavior discontinuously at different instance scales. \\

  8 & V7 & 1.27\% & Reject; V7
    & Deletion, $1\rightarrow0$ ($-1$): remove the full training-set validation requirement after small tests.
    & A small sample may miss regressions elsewhere. \\

  9 & V7 & 1.01\% & Accept; V9
    & Replacement, $1\rightarrow2$ ($+1$): revisit structure after renewed stagnation and admit only mechanisms that change target weak-region search behavior.
    & After modest gains, a renewed plateau calls for a clear trigger to reconsider the architecture. \\

  10 & V9 & 1.25\% & Reject; V9
    & Addition, $0\rightarrow1$ ($+1$): stop full-budget evaluation of costly mechanisms after no small-sample average gain.
    & Early screening may exclude costly mechanisms that benefit weak regions. \\
\end{longtable}
\endgroup

\ifdefined\TEVCStandaloneSupplement
  \clearpage
\else
  \Needspace{5\baselineskip}
\fi
\section*{Artifact Availability and Execution Templates}
\label{app:skills}

\subsection*{Anonymous Skill Artifacts}
\label{app:skill-artifacts}

The anonymous repository for the complete Initial Skill, SkillOpt-evolved solving skill, the final solving skill evolved by AutoSECO, and fixed evolver skill is \url{https://anonymous.4open.science/r/ASECO-C5D7/}.

\ifdefined\TEVCAppendicesInMain
  \Needspace{25\baselineskip}
\else
  \Needspace{8\baselineskip}
\fi
\subsection*{Problem-Independent Execution Templates}
\label{app:execution-templates}

The five templates show the shared task-definition, configuration, and prompt structure. Angle-bracketed placeholders are replaced with task-specific values or file locations before execution. The solver configuration supplies training inputs, including optional reference solutions, while the evolution configuration additionally identifies evaluation inputs for the skill evolver. The prompts load external skill documents and configurations. The templates describe how these inputs are organized, while the solving and evolver skills provide the behavioral guidance.

% One stacked, full-width block per template: this is the layout of the
% original supplementary PDF and is now used by both editions. The former
% two-up minipage variant is kept behind \TEVCTwoUpTemplates for reference.
\ifdefined\TEVCTwoUpTemplates
\par\noindent\begin{minipage}[t]{0.48\linewidth}\vspace{0pt}
\paragraph{Solver-run configuration}
\VerbatimInput[fontsize=\normalsize,breaklines=true,breakanywhere=true,frame=single,framesep=2mm,breaksymbolleft={},breaksymbolright={}]{templates_solver_run_configuration.txt}
\end{minipage}\hfill
\begin{minipage}[t]{0.49\linewidth}\vspace{0pt}
\paragraph{Skill-evolution configuration}
\VerbatimInput[fontsize=\normalsize,breaklines=true,breakanywhere=true,frame=single,framesep=2mm,breaksymbolleft={},breaksymbolright={}]{templates_skill_evolution_configuration.txt}
\end{minipage}\par\smallskip

\ifdefined\TEVCAppendicesInMain
  \vspace{2\baselineskip}
\fi
\paragraph{Problem-definition interface}
\VerbatimInput[fontsize=\normalsize,breaklines=true,breakanywhere=true,frame=single,framesep=2mm,breaksymbolleft={},breaksymbolright={}]{templates_problem_definition.txt}

\par\noindent\begin{minipage}[t]{0.49\linewidth}\vspace{0pt}
\paragraph{Solver-run prompt}
\VerbatimInput[fontsize=\normalsize,breaklines=true,breakanywhere=true,frame=single,framesep=2mm,breaksymbolleft={},breaksymbolright={}]{templates_solver_run_prompt.txt}

\end{minipage}\hfill
\begin{minipage}[t]{0.49\linewidth}\vspace{0pt}
\paragraph{Skill-evolution prompt}
\VerbatimInput[fontsize=\normalsize,breaklines=true,breakanywhere=true,frame=single,framesep=2mm,breaksymbolleft={},breaksymbolright={}]{templates_skill_evolution_prompt.txt}
\end{minipage}\par\smallskip
\else
  % Stacked, full-width layout for the five execution templates, used by the
% single-PDF appendix (main.tex) and by the standalone supplementary.tex.
% The template text is inlined here so the build does not depend on separate
% .txt files (arXiv uploads have dropped them before); the originals in
% templates/*.txt remain the editable source of the same text.

\par\noindent\begin{minipage}{\linewidth}
\noindent\textbf{Solver-run configuration}\par\nobreak\smallskip
\begin{Verbatim}[fontsize=\normalsize,baselinestretch=0.92,breaklines=true,breakanywhere=false,frame=single,framerule=0.3pt,rulecolor=\color{black!55},framesep=1.5mm,breaksymbolleft={},breaksymbolright={}]
{
  "problem": "<PROBLEM_DEFINITION>",
  "train": {
    "instances": "<TRAIN_INSTANCES>",
    "targets": "<TRAIN_TARGETS>",
    "references": "<OPTIONAL_TRAIN_SOLUTIONS>"
  }
}
\end{Verbatim}
\end{minipage}\par\medskip\vspace{2pt}

\par\noindent\begin{minipage}{\linewidth}
\noindent\textbf{Skill-evolution configuration}\par\nobreak\smallskip
\begin{Verbatim}[fontsize=\normalsize,baselinestretch=0.92,breaklines=true,breakanywhere=false,frame=single,framerule=0.3pt,rulecolor=\color{black!55},framesep=1.5mm,breaksymbolleft={},breaksymbolright={}]
{
  "evolution_skill": "<EVOLVER_SKILL>",
  "initial_solver_skill": "<INITIAL_SOLVER_SKILL>",
  "solver_run_config": "<SOLVER_RUN_CONFIG>",
  "solver_run_prompt": "<SOLVER_RUN_PROMPT>",
  "problem": "<PROBLEM_DEFINITION>",
  "workspace": "<EVOLUTION_WORKSPACE>",
  "test": {
    "instances": "<EVOLUTION_INSTANCES>",
    "targets": "<EVOLUTION_TARGETS>"
  },
  "rounds": 10
}
\end{Verbatim}
\end{minipage}\par\medskip\vspace{2pt}

\par\noindent\begin{minipage}{\linewidth}
\noindent\textbf{Solver-run prompt}\par\nobreak\smallskip
\begin{Verbatim}[fontsize=\normalsize,baselinestretch=0.92,breaklines=true,breakanywhere=false,frame=single,framerule=0.3pt,rulecolor=\color{black!55},framesep=1.5mm,breaksymbolleft={},breaksymbolright={}]
Read the Solver-development Skill:

`<SOLVER_SKILL>`

Read the Solver-run configuration:

`<SOLVER_RUN_CONFIG>`

Develop, validate, and deliver the Solver according to the Skill, problem contract, and configured information boundary.
\end{Verbatim}
\end{minipage}\par\medskip\vspace{2pt}

\par\noindent\begin{minipage}{\linewidth}
\noindent\textbf{Skill-evolution prompt}\par\nobreak\smallskip
\begin{Verbatim}[fontsize=\normalsize,baselinestretch=0.92,breaklines=true,breakanywhere=false,frame=single,framerule=0.3pt,rulecolor=\color{black!55},framesep=1.5mm,breaksymbolleft={},breaksymbolright={}]
Read the evolver skill:

`<EVOLVER_SKILL>`

Read the Skill-evolution configuration:

`<SKILL_EVOLUTION_CONFIG>`

Complete the configured evolution rounds. In each round, diagnose the current Solver-quality limitation, derive a problem-independent revision direction, and evaluate the candidate Skill through a fresh Solver-development run.
\end{Verbatim}
\end{minipage}\par\medskip

% The complete problem-definition interface stays together on a fresh page.
\clearpage
\noindent\begin{minipage}{\linewidth}
\textbf{Problem-definition interface}\par\nobreak\smallskip
\begin{Verbatim}[fontsize=\normalsize,baselinestretch=0.92,breaklines=true,breakanywhere=false,frame=single,framerule=0.3pt,rulecolor=\color{black!55},framesep=1.5mm,breaksymbolleft={},breaksymbolright={}]
---
id: <PROBLEM_ID>
name: <PROBLEM_NAME>
objective: <MINIMIZE_OR_MAXIMIZE>
quality_field: <QUALITY_METRIC>
instance_extensions: [<INSTANCE_EXTENSION>]
---

# Problem Definition

State the optimization problem in mathematical or unambiguous operational terms. Define the decision variables, feasibility conditions, and objective without referring to a particular dataset or directory.

# Evaluation Domain

Describe the instance family covered by the evaluation. Specify supported sizes, data types, conventions, and excluded variants. The description must be sufficient to determine whether an instance belongs to the intended domain.

# Objective and Evaluation Rules

Define whether lower or higher values are better. State how the evaluator validates feasibility, recomputes the objective, and derives the reported quality metric.

```text
<QUALITY_METRIC_DEFINITION>
```

Explain any reference values, aggregation rules, tolerances, or tie-breaking rules used by the evaluator.

# Input Contract

Describe the instance format and every field that the Solver may read. Identify stable instance IDs and any numerical conventions required for parsing.

# Output Contract

Provide the exact machine-readable output schema. Replace the example below with the fields required by the problem.

```json
{
  "instance": "<INSTANCE_ID>",
  "solution": "<PROBLEM_SPECIFIC_SOLUTION>"
}
```

State the type, range, and feasibility requirements of every field. List all prohibited fields or auxiliary outputs that the evaluator rejects.

# Constraints and Resource Limits

Specify the allowed language and dependencies, memory limit, per-instance time limit, parallelism policy, and any prohibited external tools. If budgets vary by instance size, define the complete size-to-budget mapping here.
\end{Verbatim}
\end{minipage}\par

\fi

  \par
  \endgroup
\fi
\end{document}